%% file: fulltext.tex
\documentclass[letterpaper, 10 pt, conference]{ieeeconf}  
\IEEEoverridecommandlockouts
\usepackage[utf8]{inputenc}

\usepackage{amsmath,amssymb,amsfonts,amsthm}
\usepackage{bm}
\usepackage{url}
\usepackage{graphicx,graphics}
\usepackage{multirow}
\usepackage{booktabs}
\usepackage{multirow}
\usepackage[ruled,linesnumbered]{algorithm2e}
\usepackage{textcomp}
\usepackage{tabularx}
\usepackage{relsize}
\usepackage[flushleft]{threeparttable}
\usepackage{cite}
\usepackage[table,dvipsnames]{xcolor}
\usepackage{colortbl}
\usepackage{enumerate}
\let \labelindent \relax 
\usepackage[inline]{enumitem}

\makeatletter
\let\NAT@parse\undefined
\makeatother
\usepackage{hyperref}  
\hypersetup{colorlinks=true, citecolor=Red, linkcolor=black, urlcolor=black}

\input{macros}

\renewcommand{\baselinestretch}{0.993}

\title{\LARGE \bf %
GIFT: Geometry-Induced Functional Transfer for Category-level Object Manipulation 
}
\author{Cristiana de Farias$^{1*}$ ~ Luis Figueredo$^{2,3}$ ~ Riddhiman Laha$^{2}$ ~ Maxime Adjigble$^{1}$\\ Brahim Tamadazte$^{4}$ ~ Rustam Stolkin$^{1}$ ~ Sami Haddadin$^{2}$ ~ Naresh Marturi$^{1}$
\thanks{This work was supported by the UK National Centre for Nuclear Robotics, and by Horizon Europe project REBELION grant 101104241. $^{1}$Extreme Robotics Laboratory, School of Metallurgy and Materials, University of Birmingham, Birmingham, United Kingdom.$^{2}$Munich Institute of Robotics \& Machine Intelligence, Technische Universität München (TUM), Germany.  $^{3}$School of Computer Science, University of Nottingham, UK. $^{4}$Sorbonne Universit\'e, ISIR, Paris, France.  $^{\dagger}$Corresponding Author: {\tt CXM1029@alumni.bham.ac.uk}.}
}

\begin{document}\sloppy
\maketitle
\thispagestyle{empty}
\pagestyle{empty}
\bstctlcite{IEEEexample:BSTcontrol}
\maketitle
\input{sections/abstract.tex}
%
\section{Introduction}
\label{sec:intro}

\input{sections/introduction.tex}
%
\section{Related Work}
\label{sec:literature}
\input{sections/relatedwork}

\section{Problem description and preliminaries}
\label{sec:prelim}
\input{sections/problem}
%
%
\section{Methodology}
\input{sections/method}
%
\section{Experimental validations}
\label{sec:experiments}
\input{sections/experiments}
%
%
\section{Conclusion}
\label{sec:conclusion}
\input{sections/conclusion.tex}
\typeout{}
\bibliographystyle{IEEEtran}
\bibliography{references}

\end{document}

%% file: macros.tex
\def\dq#1{\underline{\boldsymbol{#1}}}

\def\dualquatset{\mathbb{H}\otimes\mathbb{D}}

\theoremstyle{definition}
\newtheorem*{definit}{Definition}

\theoremstyle{definition}
\newtheorem*{problem}{Problem Statement}

\DeclareMathOperator*{\argmin}{arg\,min}

\newcommand{\sel}{\text{sel}}
\newcommand{\ctc}{\text{contacts}}

\newcommand{\skill}{\mathcal{T}}
\newcommand{\dem}{\text{dem}}
\newcommand{\shape}{\mathbf{\mathcal{S}}}
\newcommand{\trajectory}[1]{\mathbf{\dq{\mathcal{#1}}}}
\newcommand{\scene}{\mathbf{\Theta}}

%% file: sections/abstract.tex
\begin{abstract}
Robotic manipulation of unfamiliar objects in new environments is challenging due to limited generalisation capabilities. We propose a new skill transfer framework, GIFT (Geometry-Induced
Functional Transfer), which enables a robot to transfer complex object manipulation skills and constraints from a single human demonstration. Our approach addresses the challenge of skill acquisition and task execution by deriving geometric representations from demonstrations focusing on object-centric interactions. By leveraging the Functional Maps (FMC) framework, we efficiently map interaction functions between objects and their environments, allowing the robot to replicate task operations across objects of similar topologies or categories, even when they have significantly different shapes. Additionally, our method incorporates screw interpolation (ScLERP) for generating smooth, geometrically-aware robot paths to ensure the transferred skills adhere to the demonstrated task constraints. We validate the effectiveness and adaptability of our approach through extensive experiments, demonstrating successful skill transfer and task execution in diverse real-world environments without requiring additional training.
\end{abstract}

%% file: sections/introduction.tex
Robotic autonomy in human-centred settings rests on the robot’s capability to perform complex tasks in diverse, unknown, or partially known environments. Success often hinges on preserving task-defining constraints, contact relationships, relative poses, and motion primitives, rather than merely reproducing motions or joint trajectories. This topic has recently gained traction thanks to advances in vision-language-action and generative policies that enable zero-/few-shot capabilities by scaling pre-trained data and model size to open-world applications. Well known examples include RT-2 and RT-X \cite{brohan2023rt2,openx2023rtx}, OpenVLA \cite{kim2024openvla}, and Octo \cite{octo2024}, which co-train on large, heterogeneous robot datasets to achieve instruction-conditioned manipulation across embodiments, a practice also explored in some model-based methods. Yet these models typically lack explicit physical and geometric priors and exhibit brittle behaviour when tasks require tight constraint satisfaction or when scenes depart from the training distribution, even as they continue to improve with scale. In parallel, generative visuomotor policies based on diffusion and, more recently, flow matching have achieved strong data efficiency and multimodal action generation from less data-intensive demonstrations, but they also tend to degrade under distribution shift and provide no guarantees that contact, pose, or kinematic constraints will be upheld at execution time \cite{chi2023diffusion,hu2024adaflow}. This gap becomes critical even in routine scenarios. For instance, a robot trained to manipulate a bottle by picking it up, stirring its contents, and placing it on a table may fail if the bottle is replaced with one of a significantly different shape. Even if manually guided to a new pose, trajectory transfer can still fail (without larger training data) because of changes in the scene and reference frame. In other words, performance often collapses when the bottle’s shape or topology changes unless task-consistent interactions and their invariants are represented explicitly and transferred reliably across instances. Addressing this limitation requires new ways of encoding tasks and actions that generalise across objects and scenes while respecting the inherent constraints that define successful performance. To this end, this paper explores geometry-aware approaches that facilitate the transfer of task execution capabilities, enabling robots to adapt effectively to new scenarios, objects, and conditions even under limited data.

\begin{figure}
    \centering
    \includegraphics[width=\linewidth]{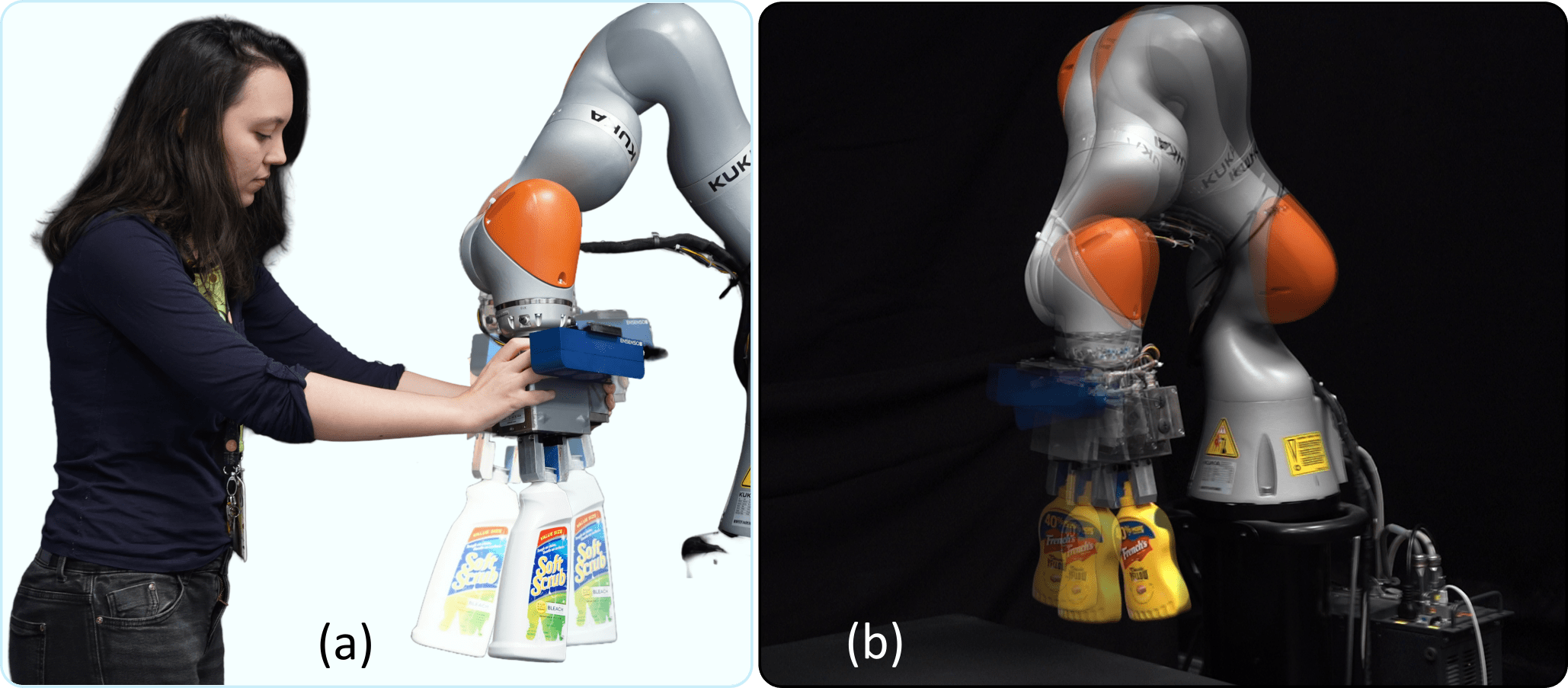}
    \caption{(a) User demonstrates the bottle shaking operation, and (b) robot imitates it on a different bottle using the proposed skill transfer framework.}
    \label{fig:bottle_shake_demonstration}
\end{figure}

We introduce a new paradigm to address this problem, casting geometric representation not as an auxiliary cue but as the backbone of skill representation and transfer. From a single kinesthetic demonstration, we derive object-centric interaction functions that encode both task-specific grasp states and the relative constraints and object-robot and object-environment interactions. These functions are defined on the object’s topology and are transported to novel, category-level instances via functional correspondences, enabling transfer to shapes with the same topology even when their geometry differs significantly. The transported interaction is then executed through a geometry-consistent path generator based on constant-screw interpolation in task space, which preserves the demonstrated relative transformations along the trajectory and blends reactively in real time. Unlike purely end-to-end pipelines, our proposed method, {GIFT}, Geometry-Induced Functional Transfer, is designed from the ground up around geometric structure to promote interpretability, strict constraint fidelity, and data efficiency. The skill can be transferred to many new instances without additional training; the constraint set is preserved by construction rather than inferred; and a single demonstration seeds an expandable family of actionable experiences via correspondence-driven projection to new shapes \cite{Farias2022FMGrasp}. 

Our formulation builds on two complementary, geometry-first insights that have remained largely disconnected in prior robotics literature. First, object geometry can drive category-level transfer of grasps and task affordances when robust correspondences are available. For instance, methods such as \cite{manuelli2019kpam,gao2021kpam,Farias2022FMGrasp,Farias2024PFMGrasp} have shown that semantically anchored geometric representations enable transferring action goals across instance motivating interaction functions defined on surfaces rather than on brittle instance templates. We extend this line by explicitly encoding robot–object and object–environment relations from demonstrations and transporting them through functional correspondences, which we instantiate via functional-map-based operators \cite{OvsjanikovFMOriginal} to handle significant shape changes while respecting topology. Second, constant-screw interpolation (ScLERP) provides a coordinate-invariant mechanism to reconstruct trajectories that conserve the implicit geometric constraints embedded in the demonstration and within the object (regardless reference-frames) \cite{allmendinger2018coordinate}, while also supporting reactive online blending and smooth tracking in novel scenes. The integration is smooth and ensures the geometric information transferred from within the category of objects, that is, to new unseen objects which might have different reference frames is respected. This is due to the known properties of screw representation, which is ad-invariant \cite{allmendinger2018coordinate,laha2022coordinate,figueredo2016kinematic}, that is, it respects changes in the reference frames in the observed objects, in contrast to decoupled methods, ubiquitously used in the literature.

Together these ideas yield GIFT, a coherent, geometry-embedded system in which (i) interaction functions formalise what must be preserved across object instances; (ii) functional correspondences instantiate where on a new shape those interactions live; and (iii) constant-screw interpolation dictates how motions evolve so that relative constraints remain invariant under changing reference frames. This design yields \textbf{explainability} (every step is anchored to surface functions and screw geometry), \textbf{predictability} (constraint satisfaction follows from representation and interpolation choices), \textbf{reactivity} (online blending under disturbances while staying on the constraint manifold), and data efficiency (one-shot transfer without policy retraining). We validate these claims in cluttered, real-world scenes using a 7-DoF platform with wrist RGB-D sensing, demonstrating consistent transfer to previously unseen instances while preserving the original geometric constraints throughout execution.

%% file: sections/relatedwork.tex
Among the many challenges related to skill transfer and learning, this work mainly focuses in two areas which are often studied separately, (i) few-shot learning of complex of complex manipulation skills; and (ii) semantic or geometric correspondence for category-level transfer. In an end-to-end approach, this problem has also been recently addressed via recent advances in (iii) generative visuomotor policies for zero-/few-shot control.

Executing tasks in new environments is often approached through imitation learning methods such as behaviour cloning or video demonstration~\cite{huang2020motion,bonardi2020learning,torabi2019imitation,liu2018imitation}. While effective in constrained settings, these approaches generally require extensive retraining and large datasets. Dynamic Movement Primitives (DMPs)~\cite{ijspeert2002movement,ijspeert2013dynamical,calinon2018learning} provided an alternative representation for encoding skills, but they are difficult to adapt to waypoint-based, object-centric interactions~\cite{paraschos2013probabilistic}. Probabilistic extensions such as ProMPs improved generalisation, yet typically converge only within the local region of the demonstration, limiting applicability to unseen scenarios~\cite{laha2022coordinate,vorndamme2022integrated}. More recently, relative-constraint approaches~\cite{laha2021point} have shown that single-shot demonstrations can enable efficient, real-time deployment with object-centric constraints; however, their scope often remains tied to the specific object instance and manually defined reference frames used during demonstration, restricting transferability.

To further advance robot generalisation, recent works have relied on semantic or geometric feature correspondences that align object instances so that task goals can be transported across shapes and objects. For example, Wen et al.~\cite{schaal_Grasp_in_Clutter} employed dense correspondence for bolt repositioning and insertion in cluttered scenes, while Tekden et al.~\cite{Tekden_yasemin_Grasp_Transfer} transferred grasps across object categories by leveraging shared semantic parts such as lids or handles. For full skill transfer of category-level objects, tasks have been described using semantic keypoints extracted from object point clouds~\cite{manuelli2019kpam,gao2021kpam}. Neural Descriptor Fields (NDFs) extend this paradigm by learning implicit object models that capture task-informed relationships between descriptors, enabling category-level generalisation for manipulation~\cite{simeonovdu2021ndf,pmlr-v205-simeonov23a}. More recently, transformer-based frameworks such as DiNoBot~\cite{di2024dinobot} have been proposed for imitation learning. Despite these advances, such methods often rely on resource-intensive (re)training, extensive manual labelling, or large black-box pre-trained models. Furthermore, planning task-oriented grasps and object-relative trajectories in unseen scenarios---where geometry and reference frames change---is particularly fragile when trajectory generation is decoupled from the semantic and geometric assumptions used for grasp or object information transfer. This is precisely the setting in which GIFT excels. It integrates correspondences from keypoints or descriptor fields and immediately enacts them via geometry-consistent screw interpolation with explicit constraint preservation, yielding closed-loop behaviour that respects object-relative invariants rather than isolated waypoints.

In contrast, a third strand scales foundation models and generative visuomotor policies to achieve broad zero-/few-shot coverage across robots and tasks. RT-2 showed that co-training VLA models on web knowledge and robot trajectories unlocks semantic manipulation from language instructions \cite{brohan2023rt2}, Open-X Embodiment and RT-X established large, standardised multi-robot datasets and policies for cross-embodiment transfer \cite{openx2023rtx}. OpenVLA provided an open VLA baseline amenable to efficient fine-tuning \cite{kim2024openvla} and Octo delivered an open generalist policy trained on hundreds of thousands of trajectories \cite{octo2024}. In parallel, diffusion policy and newer flow-matching formulations substantially improve data efficiency and multimodal action generation from limited demonstrations \cite{chi2023diffusion,hu2024adaflow,chisari2024pointcloud}. Nevertheless, these systems typically lack explicit geometric or physical priors and provide no guarantees that task-critical constraints will be upheld during execution, often degrading under shape/topology shift or camera/pose drift. Recent efforts such as ReKep, \cite{huang2024rekep} bring constraints closer to the closed-loop by continuously prompting large VLMs for keypoints and code-level relations, but they still rely on repeated perception–language inference and strong open-world semantics, which increases computational load and often leads to degradation during long-horizon control. Even widely successful platforms such as ALOHA illustrate the sensitivity of zero-shot policies to viewpoint and scene distribution, with generalisation often confined to narrow camera frustums and object placements \cite{zakka2023aloha}. These models are improving rapidly and will continue to broaden coverage. Our aim is not to compare GIFT directly to such methods, but rather to offer a fully orthogonal and complementary approach GIFT targets guaranteed, geometry-aware execution and one-shot transfer. In future work, this geometric grounding can be layered beneath VLAs or diffusion/flow policies to harden real-world behaviour by enforcing topology-anchored functional interactions and screw-consistent trajectories. We therefore do not report head-to-head scoreboards against evolving, data-hungry generalist baselines whose training assumptions and objectives differ markedly from ours, instead, we target the regime where explicit constraint fidelity, explainability, and predictable generalisation across substantial shape variation are paramount.

%% file: sections/problem.tex
\subsection{Problem definition} \label{subsec:chskill_problemstatement}
This work focuses on the challenge of generalising task execution to new category-level objects based on a single-user demonstration. To address this, we define a skill as:
\begin{definit}[Skill]
A skill is defined as the set of task operations $\skill_i$ in which every operation is comprised of trajectories and functions defined over an object's surface that encode the geometric characteristics and task-specific requirements for performing a desired manipulation task. %
It can be expressed as a tuple
\begin{equation}
\skill = \{ 
\trajectory{X},
\mathbf{f}_{\left (\text{task}\right )}  
\},
\label{eq:taskoperation}
\end{equation}
where, $\trajectory{X}=\{\dq x_0, \dq x_1, \dq x_2 \dots \dq x_n\}$ represents the robot's end-effector trajectory from the initial pose $\dq x_0$ to the final pose $\dq x_n$, and $\mathbf{f}_{\left (\text{task}\right )} \in \shape$ is an interaction function defining the task-specific relationships between the robot, the environment, and other objects in the scene, with $\shape$ being the Riemannian manifold representing an object's shape.
\end{definit}

During deployment in new scenes, it is assumed that the initial and final poses of relevant objects adhere to implicit geometric-aware task-relevant constraints obtained during a successful demonstration. These 
object-centric 
constraints,  
defined by a sequence of constant screw-connected segments, 
are captured by robot-object and object-environment interaction functions, forming a task operation tuple as in \eqref{eq:taskoperation}. These constraints are expressed as a surface function that, when coupled with FMC matching, can be transferred to new objects of the same category. Screw transformations are applied to the new key poses and segments, with screw interpolation ensuring boundary constraints between the interpolated poses.
\begin{problem}
Let a demonstration scene $\scene^{\dem}$ composed of $N$ objects be represented as
$
    \scene^{\dem} = \{ \shape^{\dem}_1, \shape^{\dem}_2, \dots, \shape^{\dem}_N \} \label{eqch04:scene_dem}
$ with  a sequence of $k$ operations kinesthetically demonstrated by the user and described by the tuple \eqref{eq:taskoperation}, i.e.,  
\begin{equation}
\skill_{{i}}^{\dem} = e\{ \trajectory{X}^\dem_i, \mathbf{f}_{\left (\text{task}, {i}\right )}\}, \quad i = 1, \ldots, k.  \label{eqch04:skill_general}
\end{equation} 
Compute the corresponding robot operations $\skill'_{i}$ that enable the robot to replicate the demonstrated skill with new objects and within the new environment, for instance, a new scene $\scene'$, containing $M$ objects. 

The aforementioned problem is outlined in two key steps:
\begin{itemize}
    \item[({\checkmark})] {\bfseries Transferability of interaction functions:} Interaction functions describing spatial and task-specific constrained relationships from the demonstrated scene $\scene^{\dem}$ must be adaptable to new objects within the same category. This adaptation should account for differences in shape and potential deformations of the new objects.
    \item[({\checkmark})] {\bfseries Maintenance of geometric constraints:} All implicit geometric constraints observed during the demonstration must be consistently maintained during task execution in new environments, regardless of changes in the object's position or the reference frame.
\end{itemize}
\end{problem}
\begin{figure*}
    \centering
    \includegraphics[width=0.9\textwidth]{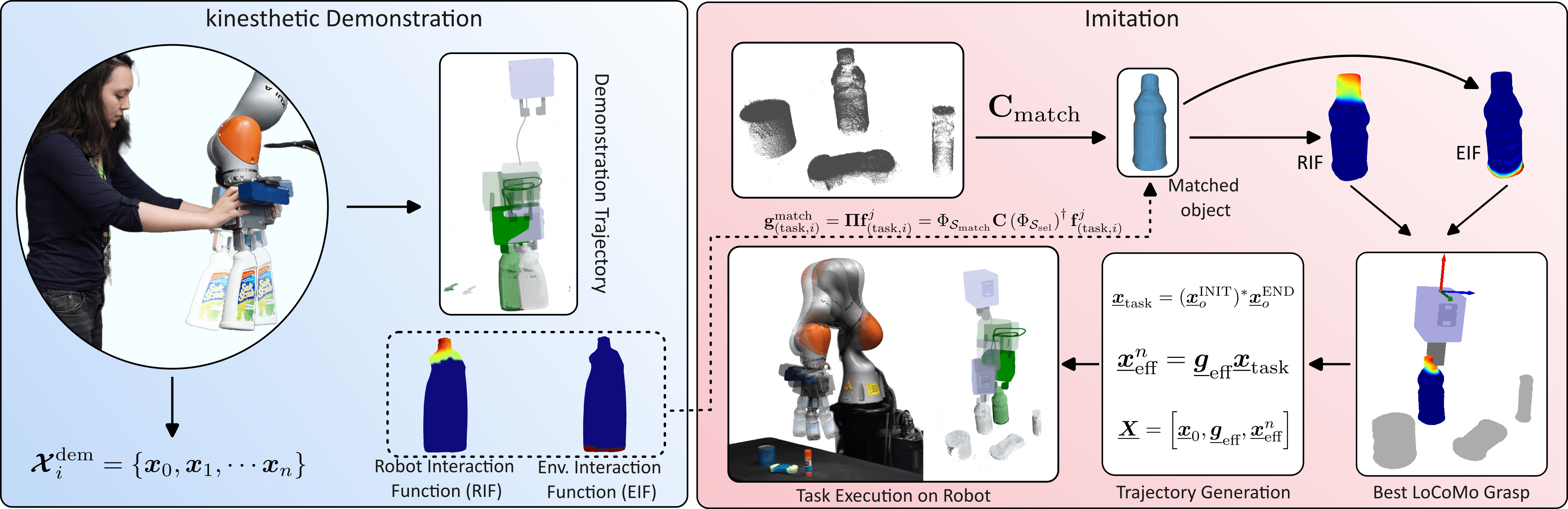}\vspace{-5pt}
    \caption{Pipeline of the proposed method (GIFT), comprising the demonstration and imitation stages. The bottle stirring skill is considered here. In the demonstration stage, kinesthetic demonstrations are performed to capture the gripper's placement and object manipulation motions. GIFT then generates functions based on these demonstrations. When the robot encounters a new scene, these functions are transferred to similar objects, allowing the robot to determine corresponding gripper poses. With a new set of waypoints and the demonstrated trajectory, the tasks are generalised to novel scenes.%
    }
    \label{fig:Pipeline}
\end{figure*} 
\subsection{Functional map correspondence}\label{Sec04:FMBasics}
%
The FMC correspondence pipeline facilitates the transfer of skill functions between objects by mapping vertices from one shape to another. For two object shapes $\mathcal{S}_1$ and $\mathcal{S}_2$, the FMC pipeline computes the map $M: \mathcal{S}_2 \rightarrow \mathcal{S}_1$. The process consists of the following steps:
\begin{enumerate}
    \item[i)] Compute a set of orthonormal bases for each shape, storing the coefficients as columns $\mathbf{\Phi}_{\mathcal{S}_1}$ and $\mathbf{\Phi}_{\mathcal{S}_2}$. Use the first $n$ eigenfunctions of the Laplace-Beltrami (LB) operator~\cite{OvsjanikovFMOriginal}. The LB basis decomposes the shape into harmonic elements, invariant to isometries and rigid motions, and is computed on 3D meshes~\cite{meyer2003discreteLBO}.
    \item[ii)] Calculate descriptor functions for each shape, which are expected to be approximately invariant across isometric shapes. Represent these functions as $\mathbf{f}_i \in \mathbf{\Phi}_{\mathcal{S}_1}$ for shape $\mathcal{S}_1$ and $\mathbf{h}_i \in \mathbf{\Phi}_{\mathcal{S}_2}$ for $\mathcal{S}_2$. Store the coefficients in matrices $\mathbf{F}$ and $\mathbf{H}$, where each column corresponds to a descriptor function. The Wave Kernel Signature~\cite{Mathieu2011WKS}, which captures intrinsic geometric properties across multiple scales, is commonly used for this purpose.
    \item[iii)] Solve the optimisation problem to obtain the optimal functional map $\mathbf{C}$ in the LB basis
    \begin{equation}
        \mathbf{C} = \argmin_{\mathbf{C}} \left( \alpha_1 E_{\text{DP}}(\mathbf{C}) + \alpha_2 E_{\text{REG}}(\mathbf{C}) \right). \label{eq:FM}
    \end{equation}
    where $\alpha_1$ and $\alpha_2$ are scalar gains. The term $E_{\text{DP}}(\mathbf{C}) = \|\mathbf{C} \mathbf{F} - \mathbf{H}\|^2$ ensures the preservation of significant shape features, while $E_{\text{REG}}(\mathbf{C})$ adds regularisation constraints for robustness~\cite{Ovsjanikov2017Tutorial}.
    \item[iv)] Refine and convert $\mathbf{C}$ into a point-to-point correspondence vector $\mathbf{T}_p$ using the \textit{ZoomOut} iterative upsampling method~\cite{melzi2019zoomout}.
\end{enumerate}
\subsection{User-guided trajectory generation}
\label{sec04:tsia}
In this work, we generates trajectories through constant-screw interpolation (ScLERP) that adhere to the geometric constraints of the demonstrated path, $\trajectory{X}^\text{DEM}$. Here, by using a screw representation, we also guarantee the ad-invariance property is transmitted. That is, even when transferring to a new object which might have a different reference frame representation, the relationship between all points in the body are maintained therefore the trajectory is consistent. In other words, it avoids the issue with decoupled methods where depending on the reference frame (observer view) the interpolated path might be different\cite{laha2021point}. This path is represented as a sequence of unit dual quaternion poses, given as
\begin{equation}
\trajectory{X}^\dem\triangleq\left\{ \begin{array}{cccc} \dq x_{0} & \dq x_{1} & \ldots & \dq x_{n}\end{array}\right\},
\end{equation}
with $\dq x_{1}, \dq x_{2}, \ldots, \dq x_{n} \in \dualquatset$. The relative path concerning the final end-effector pose $\dq x_{n}$ is then computed as
\begin{equation}
    \dq{\delta}_{i}=\dq x_{i-1}^{*}\dq x_{i},\,~~i=2,\ldots,n.
    \label{eq:dem_path_deltas}
\end{equation}

Given the desired end-effector goal pose $\dq x'_{n}$ in the new scene $\scene'$, the imitated path 
\(
    \trajectory{X}'\triangleq\left\{ \begin{array}{cccc} \dq x'_{0} & \dq x'_{1} & \ldots & \dq x'_{n}\end{array}\right\} 
\)
is obtained from \eqref{eq:dem_path_deltas}, as
\begin{equation}
    \dq x'_{i-1} = \dq x'_{n}\dq{\delta}_{i}^{*},~~i=2,\ldots,n. \label{eqch04:pathelement}
\end{equation}
This representation ensures that $\trajectory{X}'$ preserves the relative transformations of the demonstrated path throughout the imitated path, from the initial pose $\dq x'_{0}$ to the new goal $\dq x'_{n}$. %

The objective is to construct a new path starting from a different initial configuration $\dq x''_{0}$ in the task space, which does not necessarily align with the original pose $\dq x'_{0}$. For this, intermediate points are generated in the task space such that the new path $\trajectory{X}''$ blends into demonstrated path $\trajectory{X}'$, using the ScLERP method~\cite{sarker2020screw}. Given the current end-effector pose $\dq x_{\mathrm{eff}}$, at any time step, and a guiding pose $\dq x'_{i} \in \trajectory{X}'$ on the imitated path, the reference end-effector pose is
\begin{equation}
    \dq x_{r}\left( \tau \right)= \dq x_{\mathrm{eff}}\left(\dq x_{\mathrm{eff}}^{*}\dq x'_{i}\right)^{\tau}, \label{eq:imitated_path_ch03}
\end{equation}
with $\tau \in \left[0,1\right]$ being the interpolation sampling time that makes a discrete path linearly scaled along the geodesic between any current end-effector pose and the guiding pose.

%% file: sections/method.tex
The proposed skill transfer methodology is presented below.
As shown in Fig.~\ref{fig:Pipeline}, GIFT consists of two stages:
\begin{enumerate}[label={(\roman*)}]
    \item \textbf{Demonstration}, where the demonstration scene $\scene^{\dem}$ and $\skill_i^\dem$  are acquired; and
    \item \textbf{Imitation}, where the robot performs the demonstrated skill in a new scene with different objects.%
\end{enumerate}

The demonstration stage begins with the robot acquiring a complete scene $\scene^{\dem}$, by stitching point clouds from multiple scans as described in \cite{marturi2019dynamic}. Individual scene object clusters are segmented using the DBSCAN algorithm \cite{Ester_DBSCAN} and then converted to meshes using screened-Poisson reconstruction with Dirichilet boundary constraints\cite{kazhdan2013screenedPoisson}. These are stored in the database. Next, a user kinesthetically demonstrates the skill by hand-guiding the robot to perform the desired motions. This generates a set of paths, $\{\trajectory{X}_{i | i=1,2...}\}$, where each path contains poses, $\trajectory{X}_i^\dem = \{ \dq x_0, \dq x_1, \ldots, \dq x_n \}$ computed using the robot's forward kinematic model. Interaction functions (see Sec. \ref{subsec_ch04:Int_Func}), $\mathbf{f}^j_{(\text{task,i})} \in \shape^{\dem}_j$, are then computed, defining skill-specific relationships between the robot, the objects and the environment, which are crucial for generalising the skill to new scenes. Finally, $\skill_i^\dem$ can be obtained from each element in  $\trajectory{X}_i^\dem$ and $\mathbf{f}^j_{(\text{task,i})}$, as in \eqref{eqch04:skill_general}.%

In the imitation stage, the goal is to obtain an equivalent $\skill_i'$ with new functions and trajectories for a novel scene, $\scene'$. After extracting shapes from the new scene (similarly to the demonstration stage), the FMC framework is used to transfer functions $\mathbf{f}^j_{(\text{task},i)}$, resulting in
\begin{equation}
   \mathbf{g}^k_{(\text{task},i)} = \mathbf{\Pi}\mathbf{f}^j_{(\text{task},i)}, \label{eq:ch03_transfer}
\end{equation}
where $\mathbf{g}^k_{(\text{task},i)}$ represents the transferred function applied to the new object $\shape'_k$, which best matches the demonstrated object $\shape^\dem_j$. $\mathbf{\Pi}$ represents the mapping between these two shapes and their corresponding manifolds. In scenes with multiple candidate objects, a matching score (see Sec. \ref{sec:object_selection}) based on the functional map $\mathbf{C}_j$ is used to select the best match. After selecting the matching object and transferring functions, new end-effector poses are calculated, and the path is determined using ScLERP, which maintains the inherent screw-transformation-based constraints from the original demonstration.%
%
%
\subsection{Interaction functions} \label{subsec_ch04:Int_Func}
Given a scene $\scene^\dem$ and the demonstrated trajectory $\trajectory{X}_i^\dem$, we define interaction functions, $\mathbf{f}^j_{(\text{task},i)}$, over the object's surface to encode the interactions between the object, robot and the environment. Specifically, our functions include:
\begin{enumerate}[label = {(\roman*)}]
    \item Robot Interaction Functions (RIF), which define the contact regions between the robot's end-effector and the object; and
    \item Environment Interaction Functions (EIF), which define the interactions between the object and a known part of the environment.
\end{enumerate}
For demonstration scenes with multiple objects, the object of interest, $\shape_{\sel}$, is selected as the one nearest to the robot's end-effector during the last step of $\trajectory{X}_i^\dem$.
%
%
\subsubsection{Robot interaction function (RIF)}
This specifies areas on the object's surface where the robot can interact. Considering $\shape_{\sel}$ and the gripper’s kinematics, we identify ${N_{f}}$ contact points, $\mathbf{Q}^{(\ctc,f)} \in (\shape_{\sel})^{N_{f}}$, between robot's fingers and the object. The RIF is calculated as
\begin{equation}
\mathbf{f}^\text{RIF}_{(\text{task},i)} = \max \left( 1-\frac{\left\| \bar{\mathbf{Q}}^{f} - \mathbf{p} \right\|}{\lambda_D}, 0 \right)  \quad  \forall\mathbf{p} \in \shape_{\sel}
\label{eq:RIF}
\end{equation}
where $\bar{\mathbf{Q}}^{f}$ is the average of $\mathbf{Q}^{(\ctc,f)}$ for each finger, and $\lambda_D$ is a distance threshold to increase the contact region. %
\subsubsection{Environment interaction function (EIF)}
This specifies the areas of an object that interact with known elements of the environment. These elements are simplified into a set of \(N_p\) planes \(\mathbf{P} = \{ \mathcal{P}_0, \mathcal{P}_1, \dots, \mathcal{P}_{N_p} \}\), where each plane \(\mathcal{P}_k\) is defined by a point \(\mathbf{o}_k \in \mathbb{R}^3\) and a normal vector \(\mathbf{n}_k \in \mathbb{R}^3\). 

Given a selected object $\shape_{\sel}$ and the relative transformation from the demonstration, \(\dq{x}_{\text{rel}} = \dq{x}_0^* \dq{x}_n\), with \(\dq{x}_0, \dq{x}_n \in \trajectory{X}^\dem_i\) being the initial and final trajectory points, $\shape_{\sel}$ is transformed to a new configuration, updating the scene \(\scene^{\dem}\) accordingly. Once the object's pose is updated, the EIF is defined as
\begin{equation}
\mathbf{f}^\text{EIF}_{(\text{task},i)} = 
\left\{
\begin{matrix}
1, & \text{if} \quad \left \| (\mathbf{o}_k + \lambda_p \mathbf{n}_k) - \mathbf{p} \cdot  \mathbf{n}_k \right \| \leq 0 \\ 
0, & \text{otherwise}
\end{matrix}
\right.
\label{eq:EIF}
\end{equation}
$\forall \mathbf{p} \in \shape_{\sel}$. Here, \(\lambda_p\) is a distance threshold. A separate function is defined for each plane the object contacts. Typically, three planes are sufficient to determine an object's full pose.%
\subsection{Object selection}\label{sec:object_selection}
To calculate the mapping $\mathbf{C}$, an energy function is minimised to represent features in the LB bases $\mathbf{\Phi}_{\shape_1}$ and $\mathbf{\Phi}_{\shape_2}$. For perfectly isometric shapes with aligned bases, $\mathbf{C}$ is a diagonal matrix. However, with relaxed isometry or misalignments, $\mathbf{C}$ becomes diagonally dominant rather than strictly diagonal. It further degrades as the shape deviates from the reference shape. Using this property, we developed a criteria for selecting and transferring functions to the correct objects in multi-object scenes.

Given an object $\shape_{\sel}$, we find matches with all objects in
$\scene'$, resulting in a set of maps $\mathcal{C} = \{\mathbf{C}^1, \mathbf{C}^2, \dots, \mathbf{C}^j, \dots, \mathbf{C}^M\}$, where each $\mathbf{C}^j$ maps $\shape_{\sel}$ to $\shape'_j \in \scene'$. The most diagonally dominant map, $\mathbf{C}_{\text{match}} \in \mathcal{C}$ is the best match belonging to the same class. The metric $R\left(\mathbf{C}^j\right)$ is defined as
\begin{equation}
R\left(\mathbf{C}^j\right) = \left( \sum_{i,k} (1 - {w}_{ik}) |{c}_{ik}| \right) - \left( \sum_{i,k} {w}_{ik} |{c}_{ik}| \right)
\label{eq:classification_FM}
\end{equation}
where $c_{ik}$ and $w_{ik}$ are the elements of $\mathbf{C}^j$ and  the normalised weight matrix $\mathbf{W}$, respectively. Note that $\mathbf{W}$ has higher weights favouring the diagonal. This metric measures diagonal dominance, and the object in the scene with the highest score is selected as the most similar object to the demonstrated object, i.e., $\left(\shape_{\text{match}}, \mathbf{C}_{\text{match}}\right) = \underset{\shape'_j \in \scene'}{\text{arg max}} \, R\left(\mathbf{C}^j\right)$.%
%

\subsection{Skill execution in novel scenes}\label{subsec_ch04:task_execution}
%
After obtaining the demonstrated operations $\skill_i^{\dem}$, the robot is deployed in a new environment with different objects. It starts by acquiring a new scene $\scene'$, with different and previously unseen objects. Each function in $\skill_i^{\dem}$ is then transferred to the corresponding objects, i.e., given the RIF, $\mathbf{f}_{\left (\text{task}, {i}\right )} \in \shape_\sel$, a set of maps $\mathcal{C}$ is computed for every object. %
After identifying the best matching object $\shape_{\text{match}}$, the corresponding $\mathbf{C}_{\text{match}}$ is used to transfer functions to the matched shape. This transfer is computed as
\begin{equation}
     \mathbf{g}^{\text{match}}_{(\text{task},i)} = \mathbf{\Pi} \mathbf{f}^j_{(\text{task},i)} = \Phi_{\shape_{\text{match}}}\mathbf C \left ( \Phi_{\shape_{\sel}}\right )^\dag\mathbf{f}^j_{(\text{task},i)}.
\end{equation}
where $\Phi_{\shape_{\text{match}}}$ and $\Phi_{\shape_{\sel}}$ are the LB bases on the matched and demonstrated objects, respectively and $^\dag$ is the \textit{Moore–Penrose} pseudoinverse. This process is repeated for all functions in the demonstration.%
%

Typically, the defined skills result in two scenarios: 
\begin{enumerate*}[label={(\roman*)}]
    \item one RIF with two trajectories, one for approach and one for post-contact movement; and
    \item two full operations, containing a trajectory with an RIF for grasping and an EIF for skill execution.
\end{enumerate*}
In the first scenario, a simplified algorithm is applied without grasping. For skills such as pushing or button-pressing, the robot’s gripper remains closed, and a desired gripper pose is set relative to the new object's RIF. This pose serves as the goal for our ScLERP , which then interpolates a new trajectory to execute the skill. %
For scenario (ii), both the gripper’s contact regions (RIF) and the final pose with respect to the environment (EIF) are included. After the transfer, a point cloud is generated from mesh vertices where $\mathbf{g}^\text{RIF}_{(\text{task},i)} > \delta$ with $\delta$ serving as the cutoff value for the grasping region. This is then used with LoCoMo grasp planner\cite{2018_Adjigble_LoCoMo} (or alternatively \cite{Adjigble2021Spect}) to generate a ranked set of grasp poses $\mathcal{G}$. LoCoMo was selected  for its strong performance in various scenarios \cite{Bekiroglu2020Benchmark}. $\mathcal{G}$ is then filtered by removing grasps that either collide with the environment or are kinematically infeasible. Grasps are further refined by retaining only those within a cone of angle $\theta$ around the gripper’s demonstrated approach vector $\mathbf{a}_o^\dem$. 
The top-ranked grasp $\dq g^{\text{top}}$ is then selected for execution.

Once the new task-aware grasp is generated, the next step involves finding the final object and gripper poses. First, we generate point clouds $\mathbf{S}_{\sel} = \{\mathbf{p} \mid \mathbf{f}^\text{EIF}_{(\text{task},i)}(\mathbf{p}) = 1\}$ and $\mathbf{S}_{\text{match}} = \{\mathbf{p} \mid \mathbf{g}^\text{EIF}_{(\text{task},i)}(\mathbf{p}) = 1\}$, respectively for demonstrated and matched objects. Then by using point-to-point FMC, we compute the optimal rotation quaternion $\mathbf{x}_R$ to align $\mathbf{S}_{\text{match}}$ with $\mathbf{S}_{\sel}$. This alignment is further refined using Iterative closest point (ICP). 
The $\mathbf{S}_{\text{match}}$ object's transformation from its initial pose, $\dq x^{\text{INIT}}_o$ to its final pose $\dq x^{\text{END}}_o$ is given by 
\begin{equation}
    \dq x_{\text{task}} = (\dq x^{\text{INIT}}_o)^* \dq x^{\text{END}}_o.
\end{equation}
Applying this transformation to the gripper's pose in the world frame gives the final end-effector pose
\begin{equation}
    \dq x_{\text{eff}}^n = \dq g_{\text{eff}} \dq x_{\text{task}}.
\end{equation}

With the new poses, ScLERP generates new trajectories based on the demonstrated ones. We create a goals list $\dq X = \left [ \dq x_0, \dq g_{\text{eff}}, \dq x_{\text{eff}}^n \right ]$, where $\dq x_0$ represents the robot's new starting pose during imitation. For each consecutive pair in $\dq X$, a novel imitated path $\trajectory{X}_i''$ is generated. The robot then uses these paths to execute the transferred skills. 
%

%% file: sections/experiments.tex
\begin{figure}
    \centering    \includegraphics[width=0.95\columnwidth]{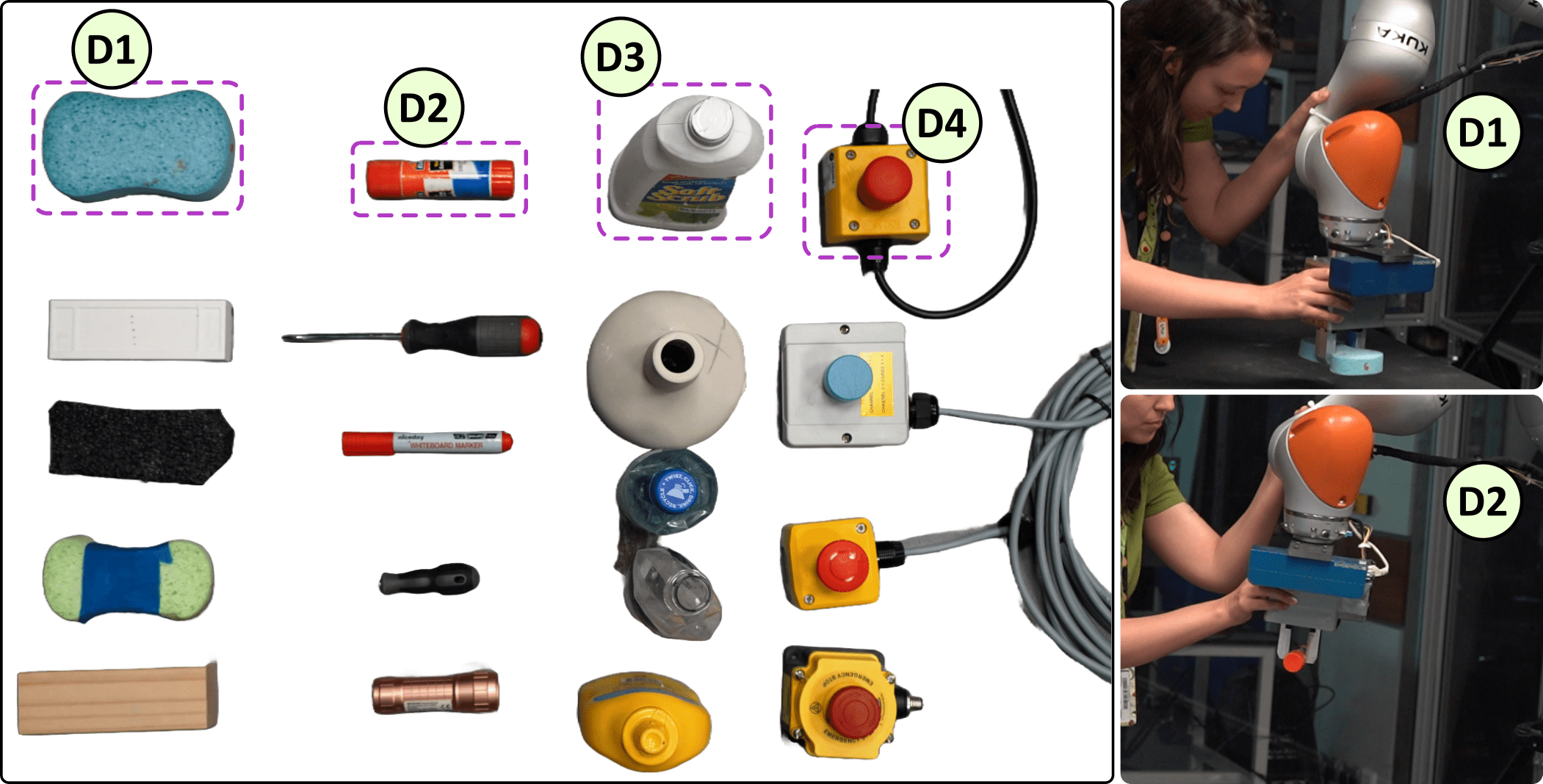}
    \caption{Dataset of objects used for experiments, with objects marked with numbers are used for demonstration. Two user demonstrations are shown.
    }
    \label{fig:data_set}
\end{figure}
\subsection{Setup description}
Our experimental setup comprises a 7-axis KUKA iiwa robot equipped with a Schunk PG 70 gripper and an Ensenso N35 3D camera mounted on the wrist. Fig. \ref{fig:data_set} illustrates the set of objects used for experiments, where objects marked with D1 -- D4 are specifically designated for skill demonstration. The skills associated with these objects are: wiping with D1, drawing a box with D2, bottle stirring with D3, and button pressing with D4. 

The FMC framework was implemented using the packages from \cite{melzi2019zoomout}, with the number of LB bases set to 85 and upsampled to 200 during refinement. All operations using dual quaternion algebra were performed with the DQrobotics package~\cite{AdornoMarinho2020}. All experiments were run on an Ubuntu 20.04 PC with an Intel i7 4-core CPU and 32 GB of RAM. The analysis began by evaluating the GIFT's ability to match shapes in a new scene with multiple objects. Following this, we assessed function transfer, focusing on RIF and EIF. Finally, we evaluated the overall skill transfer framework by examining the robot's ability to transfer and execute complete tasks in the new scene.
\subsection{Multi-object scene}
\begin{figure}
    \centering
    \includegraphics[width=0.9\linewidth]{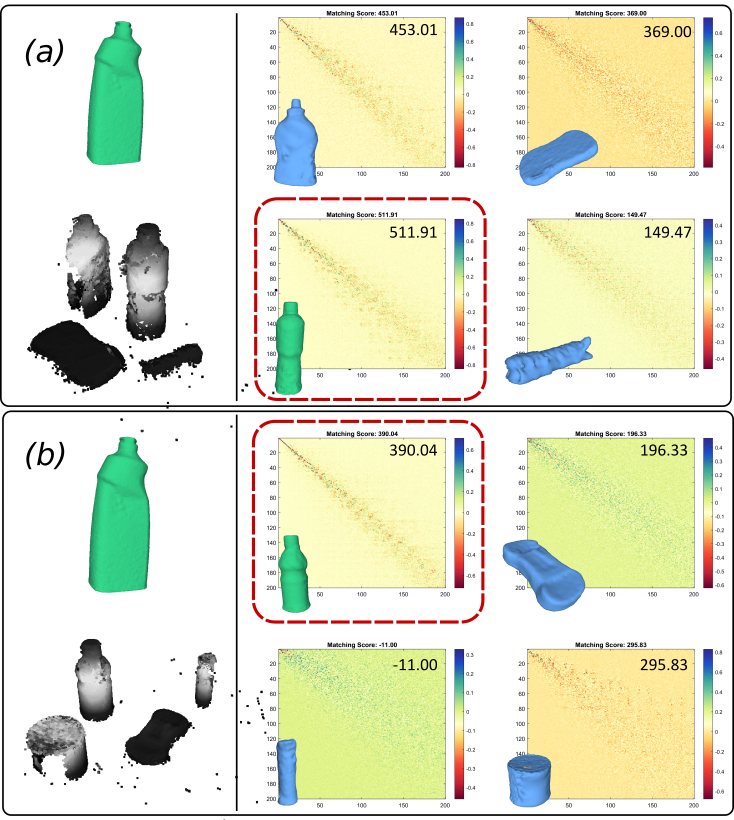}
    \caption{Illustration of object matching in two multi-object scenes. For each scene object, a map is computed and shown on the right with scores overlaid. The object with the highest matching score is selected.
    }
    \label{fig:Multi_object}
\end{figure}

We first demonstrate GIFT's ability to handle complex scenes with multiple objects. As detailed in Sec. \ref{sec:object_selection}, we compute maps for all objects in the scene and assign scores to each of them. The one with the highest score is identified as the best match. Fig.~\ref{fig:Multi_object} illustrates the matching and selection process for two different scenes, where the green mesh (demonstrated object) on the top left side is matched with all objects in the point cloud shown on the bottom left side. The computed maps are shown on the right with the corresponding matching scores overlaid. In both examples, GIFT correctly identified the best-matching object (marked with a red dotted line) that has the most diagonally dominant map. Specifically, in scene (a), we presented the robot with two different bottles in the scene. While both bottle maps were diagonal, the chosen bottle achieved the highest score, i.e., it was more similar in size and shape to the source mesh. This similarity better preserved the isometry constraints, resulting in a more optimal map. 

\subsection{Function transfer analysis}
\begin{figure}
    \centering
    \includegraphics[width=0.9\linewidth]{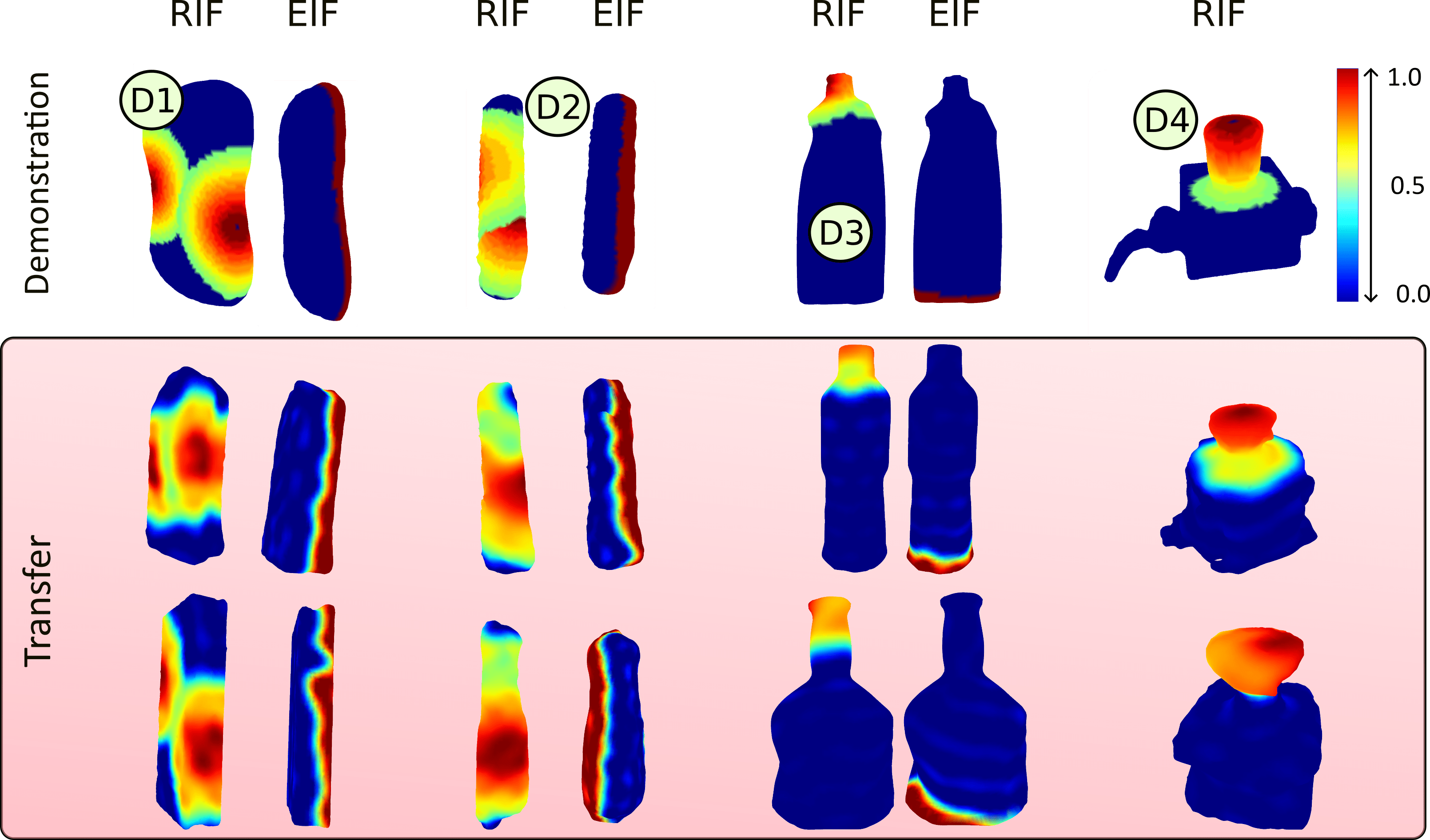}
    \caption{Illustration of function transfer from the demonstration objects (top row) to other objects within the same category.
    }
    \label{fig:function_transfer}
\end{figure}

To assess our GIFT's effectiveness in transferring the RIF and EIF, we measured the Mean Absolute Error (MAE) and standard deviation (STD) of the function transfer relative to a manually annotated ground truth, as in \cite{Farias2024PFMGrasp}. Table \ref{tab:func_transfer} presents the average results from four trials for each skill, where both RIF and EIF were transferred from the objects used for demonstration to similar category-level objects (see Fig.\ref{fig:data_set}). 
%
\begin{table}
\caption{Evaluation of function transfer across various skills.}
\centering
\begin{threeparttable}
\begin{tabular}{l|cc|cc|cc|c}
\toprule
\rowcolor{LimeGreen!35}
\textbf{Skill} & \multicolumn{2}{c|}{\textbf{RIF}} & \multicolumn{2}{c|}{\textbf{EIF}} & \multicolumn{2}{c|}{\textbf{Total}} & \textbf{Time} \\ 
\rowcolor{LimeGreen!35}
               & \textbf{MAE}\tnote{*} & \textbf{STD}\tnote{*} & \textbf{MAE}\tnote{*} & \textbf{STD}\tnote{*} & \textbf{MAE}\tnote{*} & \textbf{STD}\tnote{*} & \textbf{(s)} \\ 
\midrule
D1             & 0.175 & 0.200 & 0.201 & 0.330 & 0.188 & 0.265 & 4.16 \\
D2             & 0.159 & 0.198 & 0.170 & 0.215 & 0.165 & 0.206 & 7.89 \\
D3             & 0.127 & 0.181 & 0.061 & 0.161 & 0.094 & 0.171 & 5.84 \\
D4             & 0.145 & 0.193 & -      & -      & 0.145 & 0.193 & 11.89 \\
\bottomrule
\end{tabular}
\begin{tablenotes}
    \item[*] Values range from 0 to 1, with 0 being the best and 1 the worst.
    
\end{tablenotes}
\end{threeparttable}
\label{tab:func_transfer}
\end{table}
%
Fig.\ref{fig:function_transfer} displays the transferred RIF and EIF for all objects. We observe that the total error remains low for all skills. However, for D1, the error is slightly higher, likely due to the FMC not accounting for the object's symmetry, occasionally resulting in the rotation of the map around the object's length. Since the RIF is slightly slanted, this rotation contributes to the increased error. For less symmetric object-function pairs, such as D3 and D4, this is reduced. Nevertheless, the error remains small, and the functions are sufficiently similar to enable skill transfer. Furthermore, we analysed the time taken to transfer the functions, as FMC calculation is the most resource-intensive step in our method. Although processing time can vary based on the number of vertices and the object's geometry, the average time across all object categories is $7.4\mathrm{s}$. 

\begin{figure}
    \centering
    \includegraphics[width=0.9\linewidth]{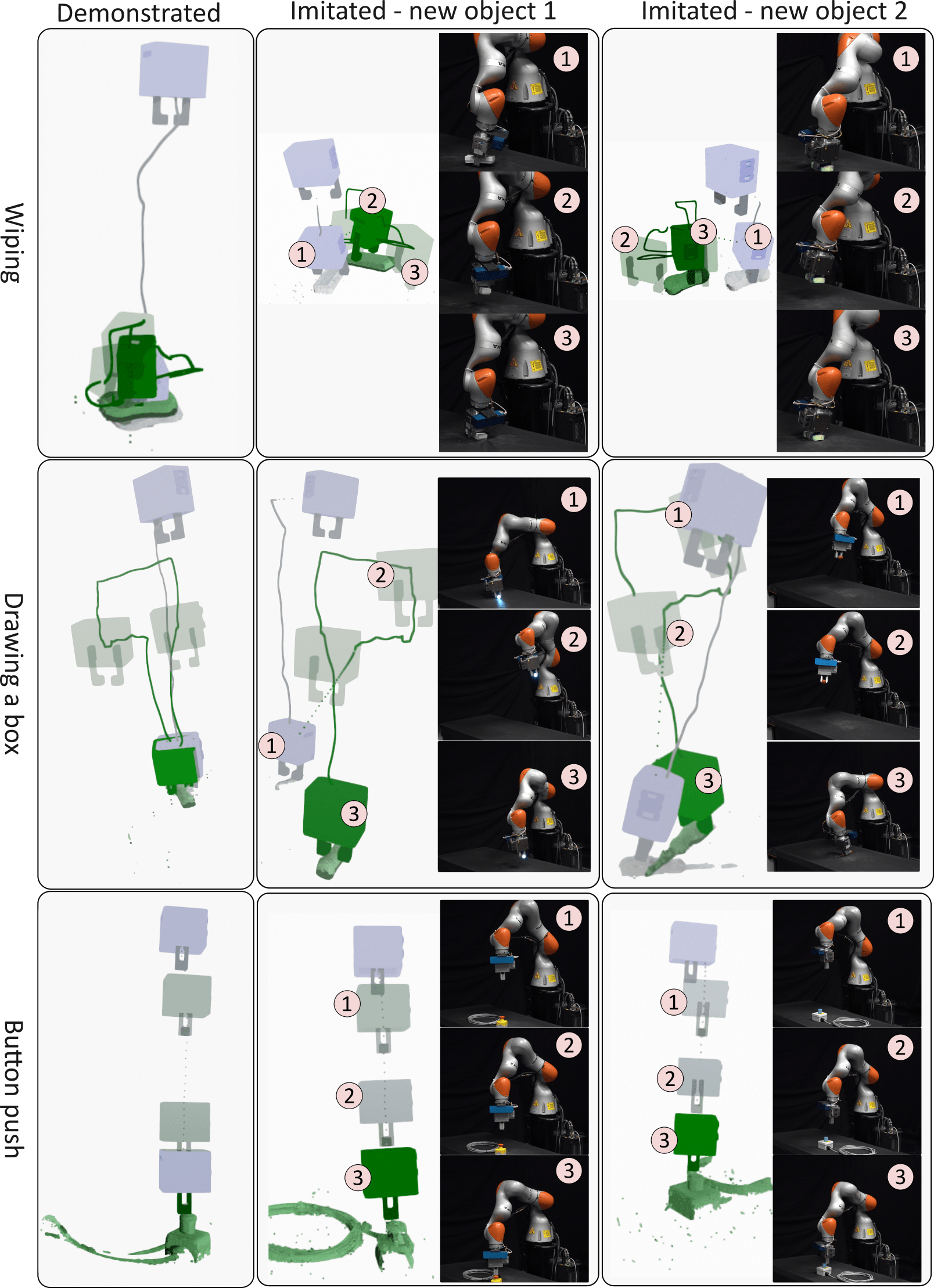}
    \caption{Illustration of three full skill transfers: wiping, drawing a box, and pressing a button, using two test objects from the dataset. (Left) Demonstrated trajectory, (middle), and (right) robot imitations in two different scenes. See the supplementary video for detailed results.}
    \label{fig:task_transfer}
\end{figure}

\subsection{Skill transfer analysis}
This section evaluates the full skill transfer performance of our approach. Fig. \ref{fig:task_transfer} shows results for three skills: wiping a surface (D1), drawing a box (D2), and button press (D4). Both demonstrated and imitated trajectories are displayed, along with screenshots of the robot during imitation. It can be seen that trajectory constraints are maintained from demonstration, and even when the final pose differs from the initial configuration, the robot follows the imitated path. The robot occasionally rotates the object, as observed with the screwdriver in the drawing skill last column, due to the FMC not accounting for object symmetry allowing alignment in either direction. Note that this is not considered a failure, though future work could include symmetry analysis to improve the target end configuration. Table \ref{tab:grasp_success} presents the average success rate, which also looks for stably grasping/touching the correct region of the object, and the average skill imitation time across four trials. All tasks were successfully accomplished with every object in our dataset, except for one failure: during a wiping trial (D1), a misalignment while grasping the wooden block (last row in Fig.~\ref{fig:data_set}) caused the gripper to collide with the edge, rotating the block.
\begin{table}
\caption{Performance analysis of various skills.}\label{tab:grasp_success}
\centering
\begin{tabular}{l|c|c}
\toprule
\rowcolor{LimeGreen!35}
\textbf{Skill} & \textbf{Success rate (\%)} & \textbf{Imitation time $\;\pm\;$ STD (s)} \\ 
\midrule
D1 & 75.0 & 166.75$\;\pm\;$4.72\\ 
D2 & 100 & 115.00$\;\pm\;$4.24 \\ 
D3 & 100 & 79.00$\;\pm\;$9.88 \\ 
D4 & 100 & 35.22$\;\pm\;$3.89 \\
\bottomrule
\end{tabular}
\end{table}

%% file: sections/conclusion.tex
In this paper, we demonstrated GIFT, a one-shot skill transfer method that effectively generalises to unseen category-level objects. Beyond enabling efficient transfer from a single demonstration, our approach highlights the applicability of embedding geometric priors directly into the representation of manipulation skills. This design choice provides stronger guarantees of constraint fidelity and interpretability compared to purely end-to-end methods. By leveraging functional map correspondences and screw-based trajectory generation, GIFT achieves reliable performance in multi-object scenes while remaining computationally efficient and data-light.
For future works, we aim to integrate GIFT with large-scale visuomotor foundation models to further bridge the gap between robust geometric grounding and the semantic flexibility of data-driven policies. 